\documentclass[runningheads]{llncs}

\usepackage{eccv}

\usepackage{eccvabbrv}

\usepackage{graphicx}

\usepackage{booktabs}

\usepackage[accsupp]{axessibility}  

\usepackage[pagebackref,breaklinks,colorlinks,citecolor=eccvblue]{hyperref}

\usepackage{orcidlink}
\usepackage{makecell}
\usepackage{multirow}
\usepackage{colortbl}
\usepackage{xcolor}
\begin{document}

\title{Revisiting Shape from Polarization in the Era of Vision Foundation Models} 


\author{Chenhao Li \and Taishi Ono \and Takeshi Uemori \and Yusuke Moriuchi}

\authorrunning{Li et al.}

\institute{
Sony Semiconductor Solutions Corporation
\email{\{Chenhao.Li,Taishi.Ono,Takeshi.Uemori,Yusuke.Moriuchi\}@sony.com}}

\maketitle

\begin{figure}[t]
  \centering
  \includegraphics[width=\linewidth]{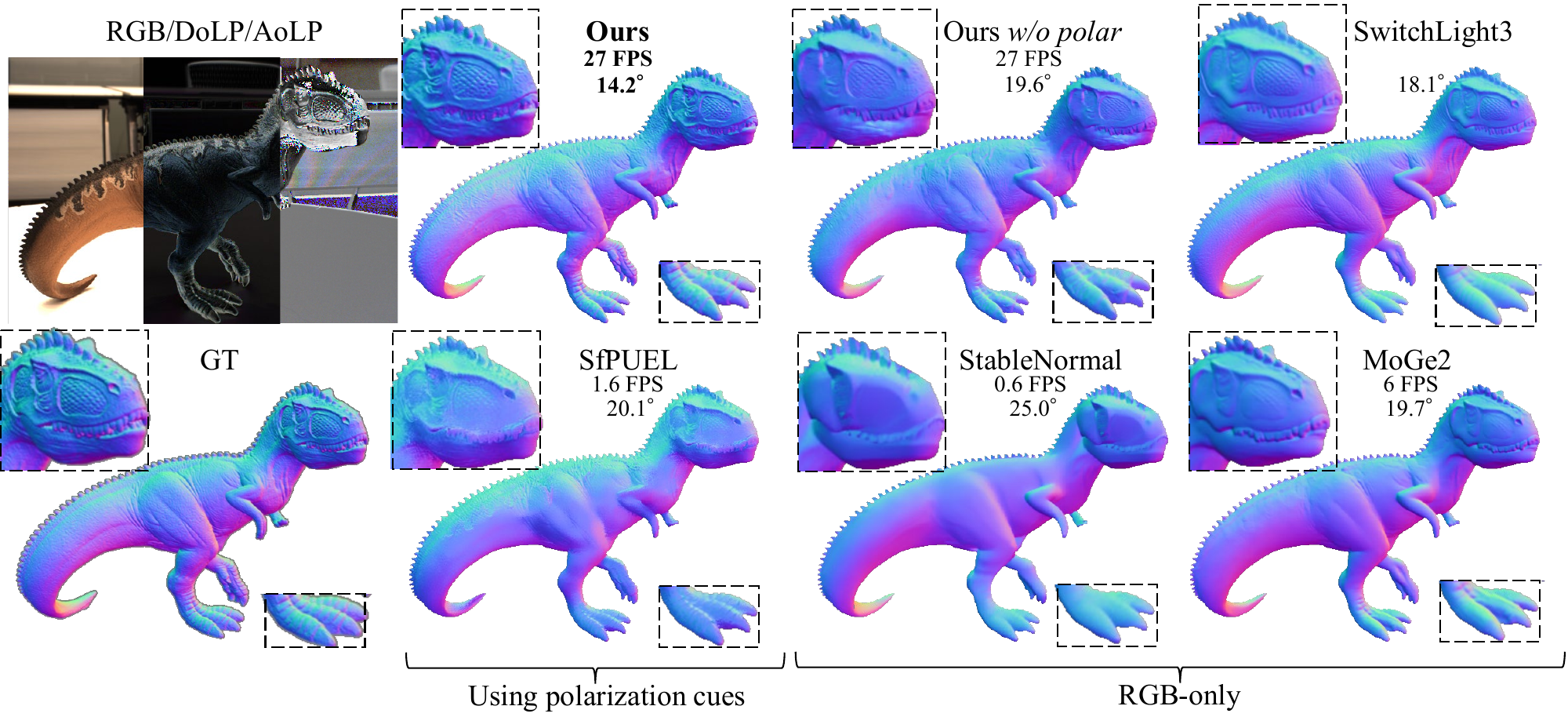}
  \caption{Our method surpasses the previous best SfP approach (SfPUEL \cite{lyu2024sfpuel}), a leading discriminative VFM (MoGe2 \cite{wang2025moge2}), a generative VFM (StableNormal \cite{ye2024stablenormal}), and a commercial inverse rendering tool (SwitchLight3 \cite{beeble_switchlight3}). Moreover, the benefit of using polarization cues is clear by comparing with our RGB-only ablation. The numbers shown below each method indicate frames per second (FPS) and mean angular error (MAE). Inference speed for all models is tested on a V100 GPU with a resolution of 512 × 612 and FP16 precision. \protect\footnotemark}   
  \label{fig:teaser}
\end{figure}

\begin{abstract}
We show that, with polarization cues, a lightweight model trained on a small dataset can outperform RGB-only vision foundation models (VFMs) in single-shot object-level surface normal estimation.
Shape from polarization (SfP) has long been studied due to the strong physical relationship between polarization and surface geometry. 
Meanwhile, driven by scaling laws, RGB-only VFMs trained on large datasets have recently achieved impressive performance and surpassed existing SfP methods. 
This situation raises questions about the necessity of polarization cues, which require specialized hardware and have limited training data.
We argue that the weaker performance of prior SfP methods does not come from the polarization modality itself, but from domain gaps. These domain gaps mainly arise from two sources. First, existing synthetic datasets use limited and unrealistic 3D objects, with simple geometry and random texture maps that do not match the underlying shapes. Second, real-world polarization signals are often affected by sensor noise, which is not well modeled during training.
To address the first issue, we render a high-quality polarization dataset using 1,954 3D-scanned real-world objects. We further incorporate pretrained DINOv3 priors to improve generalization to unseen objects. To address the second issue, we introduce polarization sensor-aware data augmentation that better reflects real-world conditions.
With only 40K training scenes, our method significantly outperforms both state-of-the-art SfP approaches and RGB-only VFMs. Extensive experiments show that polarization cues enable a 33× reduction in training data or an 8× reduction in model parameters, while still achieving better performance than RGB-only counterparts.

\end{abstract}

\section{Introduction}
\label{sec:intro}
A normal map provides a 2.5D representation of surface geometry and has been extensively studied in both computer vision and computer graphics. Accurate estimation of normal maps from single 2D images is crucial for many downstream applications, including AR, VR, robotics, and industrial inspection. However, this task is inherently ambiguous, as similar visual appearances can result from different combinations of lighting, material properties, and geometry. Traditional models are physics-based, but they heavily rely on multi-view \cite{hartley2003multiple,snavely2008modeling} or multi-light \cite{woodham1980photometric} observations to reduce ambiguity. In single-shot settings, these methods face fundamental limitations, making learning-based approaches popular solutions for this ill-posed problem.
\footnotetext{We used a commercially available Schleich dinosaur figurine purchased by the authors. No endorsement by the manufacturer is implied.}

Recent progress in single-shot surface normal estimation is mainly dominated by vision foundation models (VFMs). Existing VFMs can be roughly divided into two groups: discriminative and generative models. Although they follow different paradigms, both groups predict high-quality surface normals. Discriminative methods, such as MoGe \cite{wang2025moge, wang2025moge2}, directly map a 2D RGB image to 3D geometry using neural networks. However, their performance relies on millions of training data, leading to high training and data collection costs. Generative methods use priors from diffusion models, reducing the required scale of training data. Representative works like StableNormal \cite{ye2024stablenormal} and Marigold \cite{ke2024repurposing} only use 250K and 74K samples. However, their inference involves multiple diffusion steps, making these methods slow and unsuitable for real-time use. Overall, existing VFMs improve accuracy but remain either data-hungry or computationally expensive, motivating the search for more efficient alternatives.

Polarization, as one of the properties of electromagnetic waves, has long been studied for normal estimation due to its strong link to surface geometry. This task is known as shape from polarization (SfP). Early SfP methods \cite{ngo2015shape, cui2017polarimetric, kadambi2015polarized, riviere2017polarization, fukao2021polarimetric, ichikawa2024spiders} are mostly rule-based and recover surface normals using physical laws from observed polarization signals. To avoid well-known issues such as diffuse–specular ambiguity and $\pi$ ambiguity, these methods often rely on strict assumptions about lighting and surface materials, which limits their use in real scenes. Later, learning-based approaches \cite{ba2020deep, lyu2024sfpuel} combine polarization cues with RGB images and use neural networks to predict surface normals in a data-driven manner. Compared to rule-based methods, these approaches work under fewer assumptions and achieve higher accuracy. Moreover, under similar training data scales, prior work has shown that using RGB + polarization consistently produces more accurate surface normals than using RGB alone.

Despite the progress of SfP methods, they still perform much worse than VFMs. This situation is unexpected, given the rich geometry-related information provided by polarization cues. Therefore, a systematic re-examination of existing SfP methods is desired. We argue that the poor performance is not due to the polarization modality itself, but to several domain gaps. We highlight two main factors.
First, the training data lacks diversity and realism. SfP methods using real datasets \cite{ba2020deep} usually contain only a few hundred scenes, which is insufficient for modern deep networks. Synthetic datasets \cite{lyu2024sfpuel} are larger (about 20K scenes) but are rendered using only around 200 objects, leading to limited diversity. Moreover, due to the lack of geometry-consistent textures, a random one is usually applied, resulting in unrealistic images.
Second, noise in real polarization sensors is not well modeled. Synthetic data are clean due to an ideal camera model, while real sensors suffer from degradations, such as shot noise and lens blur. Although such noise has less effect on RGB images, it severely degrades polarization signals, especially the noise-sensitive angle of polarization.

This work focuses on addressing two key issues of existing SfP methods mentioned before. To mitigate the insufficient diversity and realism in training data, we render 40K polarized scenes using 1,954 scanned 3D objects with geometry-consistent textures from the Digital Twin Catalog \cite{dong2025digital}, and name the resulting dataset DTC-p. Despite the large number of objects in DTC-p, the performance degrades significantly when the model is applied to unseen content such as transparent objects. To improve generalization, we further integrate features from the recently pretrained model DINOv3 \cite{simeoni2025dinov3} as a prior into our model. To address the noise problem in real-world polarization signals, we introduce a polarization sensor-aware data augmentation strategy, which adds random noise and blur, and quantize the polarization images during training. The quantitative evaluation is conducted on two public datasets \cite{chen2024pisr, lyu2024sfpuel} and our constructed real-world dataset.  Our method reduces the mean angle error by 21\% compared to the previous best SfP method \cite{lyu2024sfpuel}, and 8\% compared to the best RGB-only VFM \cite{wang2025moge2}, while using only 0.45\% of the training data. 

Existing SfP works mainly aim to demonstrate that using polarization cues outperforms the RGB-only counterparts under the assumption of comparable model size and training data. 
Instead, we provide a new perspective of the value of polarization cues: we study how much we can reduce the training data and the model size by using polarization cues. Today’s VFMs are becoming expensive mainly because both the models and the datasets keep growing. We show that combining a physics prior with deep learning is an efficient way to reduce this cost. In our ablation studies, by using polarization cues, similar performance can be achieved using only 1/33 of the training data and 1/8 of the model parameters compared to their RGB-only counterparts. Moreover, while model ablations are commonly explored in previous works, dataset ablations are rarely discussed. We address this gap by conducting a comprehensive ablation study on the 3D models and environment maps used to render the training data.
Our contributions are as follows:  
\begin{itemize}
\item Achieved a milestone in single-shot object-level normal estimation, outperforming both SfP and RGB-only VFMs by a large margin.
\item Demonstrated the role of polarization sensors in the era of VFMs, namely achieving the similar performance with much less training data and a smaller network compared to RGB-only methods.
\item Conducted extensive ablation studies on both the model and dataset to analyze the sources of performance improvement. 
\end{itemize}

\section{Related Works}
\subsection{Single-shot normal estimation}
Single-shot surface normal estimation has been dominated by learning-based methods. These methods can be broadly divided into two categories: discriminative models and generative models.

\noindent \textbf{Discriminative models}
Early methods \cite{bansal2016marr, eigen2015predicting, wang2020vplnet, do2020surface, wang2015designing} formulate surface normal estimation as a discriminative task.
Omnidata \cite{eftekhar2021omnidata} advanced this direction by enabling the creation of million-scale datasets, which made it possible to train data-hungry vision transformers \cite{ranftl2021vision}.
More recently, the state-of-the-art performance is achieved by introducing an affine-invariant point map as a new 3D representation (MoGe \cite{wang2025moge, wang2025moge2}). However, these methods rely heavily on millions of training samples, which are expensive to collect and lead to high training costs. This makes continuous updates and improvements difficult. Although some efforts have been made to reduce data requirements, such as introducing inductive biases for surface normal estimation (DSINE\cite{bae2024rethinking}), the performance degradation is still noticeable. DAViD \cite{saleh2025david} shows a small but high-quality synthetic dataset is enough for high-fidelity normal estimation, but they only focus on human faces.  

\noindent \textbf{Generative models}
Another line of work leverages priors from diffusion models \cite{ho2020denoising} for surface normal estimation. Representative works like Marigold \cite{ke2024repurposing} and GeoWizard \cite{fu2024geowizard} significantly reduce the amount of required training data. StableNormal \cite{ye2024stablenormal} further introduces a coarse-to-fine strategy to mitigate the randomness of diffusion models. However, a common limitation of these methods is that inference requires multiple diffusion steps, resulting in slow runtime. Although approaches such as single-step diffusion \cite{garcia2025fine} have been proposed to accelerate inference, fine-grained geometric details are often lost.

Our method falls into the category of discriminative models. We demonstrate that polarization cues are effective in addressing both the data-hungry nature of discriminative VFMs and the slow inference speed of generative VFMs.

\begin{table}[t]
\centering
\caption{Comparison of SfP datasets}
\label{tab:dataset_comparison}
\begin{tabular}{lcccc}
\toprule
Dataset & Image type & \#Scenes & Object type & \#Objects \\
\midrule
DeepSfP \cite{ba2020deep}        & Real        & 263  & Real-world        & 25     \\
SfPW \cite{lei2022shape}          & Real        & 522  & Real-world        & --   \\
Kondo et al. \cite{kondo2020accurate}   & Synthetic   & 44K  & Manually designed & --     \\
SfPUEL \cite{lyu2024sfpuel}        & Synthetic   & 20K  & Manually designed & 244    \\
DTC-p (Ours)   & Synthetic   & 40K  & 3D scanned        & 1,954  \\
\bottomrule
\end{tabular}
\end{table}

\subsection{Polarized vision}
Polarization is an important property of electromagnetic waves and provides information that is unique and complementary to wavelength and amplitude. Polarization cues have been widely used in many applications, including de-hazing \cite{zhou2021learning}, white balance \cite{ono2022degree}, reflection removal \cite{lei2020polarized, lyu2019reflection, yao2025polarfree,tang2024learning}, vehicle detection \cite{dong2024exploiting}, shadow removal \cite{zhou2025polarization}, material segmentation \cite{liang2022multimodal}, glass segmentation \cite{kalra2020deep, qiao2023multi}, BRDF reconstruction \cite{deschaintre2021deep}, and alpha matting \cite{enomoto2024polarmatte}. Shape estimation is one of the most common applications of polarization. In recent years, significant progress has been made in SfP methods \cite{li2024neisf, li2025neisf++, dave2022pandora, han2024nersp, qiu2025high, han2025polgs, chen2024pisr, kushida2025thermal, wu2025glossy,lincetto2025multimodalstudio} based on NeRF \cite{mildenhall2020nerf} or 3DGS \cite{kerbl20233d}. However, these methods rely on multi-view inputs. Due to space limitations, this section covers only single-shot SfP methods. We roughly divide single-shot SfP methods into physics-driven and data-driven approaches.

Single-shot normal estimation is inherently ill-posed, even when polarization cues are available. As a result, physics-driven methods usually rely on strong assumptions, such as known material properties \cite{tozza2017linear}, known lighting conditions \cite{ichikawa2021shape}, local geometry smoothness \cite{chen2022perspective}, or a near-coaxial camera–projector setup \cite{baek2018simultaneous}. Another line of research reduces the ambiguity by introducing additional devices, such as time-of-flight sensors \cite{kadambi2015polarized, baek2022all} or structured polarization projectors \cite{ichikawa2025single}.  However, these assumptions and hardware requirements greatly limit the applicability of these methods in real-world scenarios.

Due to the ill-posed nature of the problem, data-driven approaches have become a more suitable solution. DeepSfP \cite{ba2020deep} is the first work that combines polarization cues with deep neural networks for normal estimation. Since then, many extensions have been proposed, including scene-level reconstruction \cite{lei2022shape}, transparent \cite{shao2023transparent}, or translucent \cite{li2024deep} objects. To further reduce ambiguity, some methods incorporate additional cues such as shading constraint \cite{lyu2023shape}, photometric stereo priors \cite{lyu2024sfpuel}, and lidar inputs \cite{scheuble2024polarization}. Despite these efforts, their performance still lags behind recent VFMs. We identify a key reason as the insufficient diversity and realism of existing training data. To address this issue, we propose a high-quality dataset for training. A comparison between our dataset and existing ones is shown in Table \ref{tab:dataset_comparison}. In addition, we observe that polarization sensor noise is overlooked in previous work. To better model real-world conditions, we introduce a polarization sensor-aware data augmentation strategy during training.

\section{Background of Polarization}

The polarization state of light is commonly represented by Stokes vectors.
Assuming linear polarization, as in most SfP works, the Stokes vector is defined as
$\mathbf{s} = [s_0, s_1, s_2]^{\top}$,
where $s_0$ denotes the intensity, $s_1$ represents the intensity difference between horizontal and vertical polarization, and $s_2$ represents the intensity difference between $45^\circ$ and $135^\circ$ polarization.
Using a polarization camera \cite{SonyPolarSens}, four linearly polarized images $\{I_0, I_{45}, I_{90}, I_{135}\}$ can be captured in a single shot, from which the Stokes parameters are computed as
\begin{equation}
s_0 = \tfrac{1}{2}(I_0 + I_{45} + I_{90} + I_{135}),\;
s_1 = I_{0} - I_{90},\;
s_2 = I_{45} - I_{135} .
\label{equ:four2stokes}
\end{equation}
From the Stokes vector, the Degree of Linear Polarization (DoLP) and the Angle of Linear Polarization (AoLP), which are widely used in SfP due to their strong correlation with surface geometry, are computed as
\begin{equation}
\mathrm{DoLP} = \frac{\sqrt{s_1^2 + s_2^2}}{s_0}, \qquad
\mathrm{AoLP} = \tfrac{1}{2}\arctan\!\left(\frac{s_2}{s_1}\right).
\label{equ:stokes2dolpaolp}
\end{equation}

\begin{figure}[t]
  \centering
  \includegraphics[width=\linewidth]{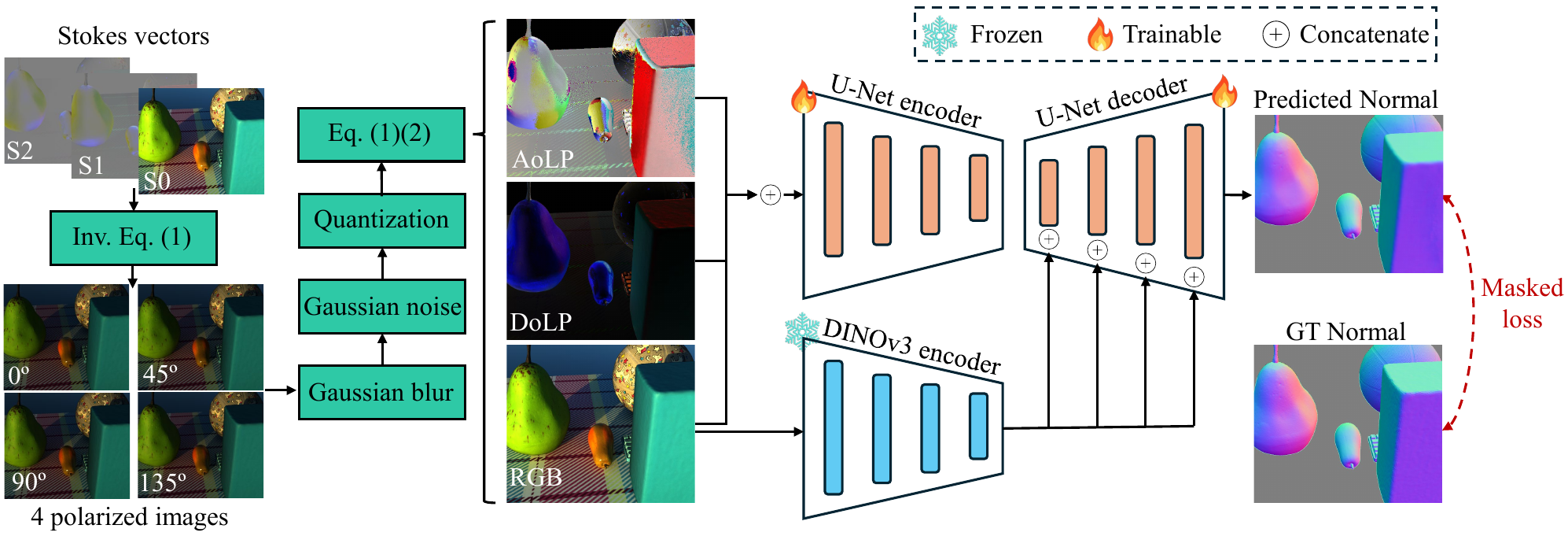}
  \caption{Data augmentation pipeline and model architecture.}
  \label{fig:model}
\end{figure}

\begin{figure}[t]
  \centering
  \includegraphics[width=\linewidth]{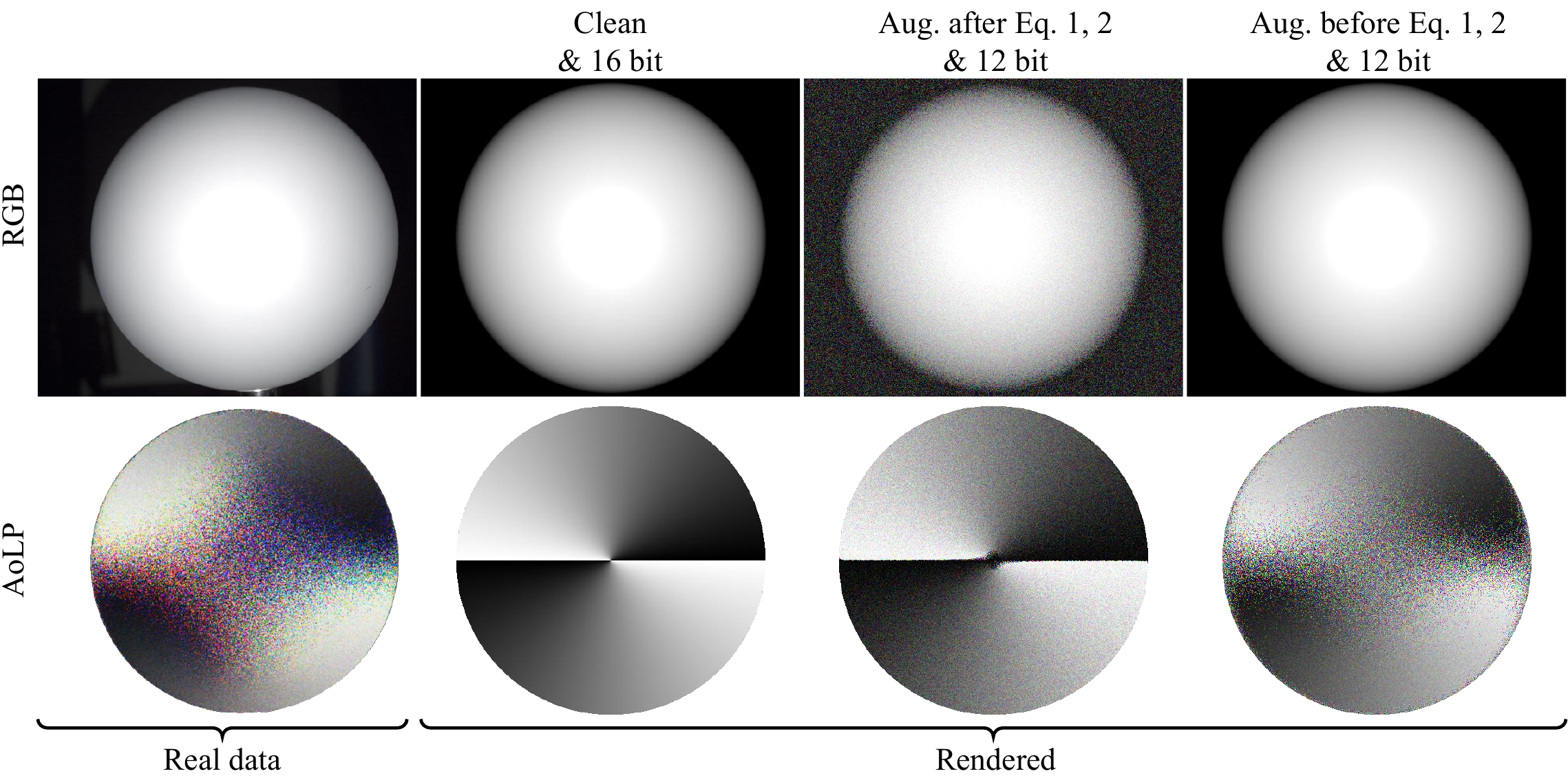}
  \caption{Visualization of a plastic ball in real and synthetic data with noise simulation. In real-world measurements, AoLP is consistently noisy due to sensor and acquisition artifacts. In contrast, rendered AoLP appears overly clean because of the idealized sensor model. Directly injecting noise into RGB or AoLP is not realistic (the noise level is amplified here for visualization). Instead, applying augmentation before polarization signal processing better matches real noise characteristics: RGB domain is less affected and AoLP noise is concentrated in regions with rapid AoLP direction changes.}
  \label{fig:augmentation}
\end{figure}

\section{Method}
We estimate surface normals from polarized images using a learning-based pipeline. The method consists of two parts: polarization sensor-aware data augmentation and an end-to-end network, including a UNet \cite{ronneberger2015u} and a pretrained DINOv3 encoder \cite{simeoni2025dinov3}. The overall architecture is illustrated in Fig. \ref{fig:model}.

\noindent \textbf{Polarization sensor-aware augmentation}
This method assumes synthetic images as training data. Therefore, careful design to alleviate synthetic-to-real domain gaps is necessary. A faithful simulation, including lens distortion, polarization extinction ratio degradation, and various sensor noises, would require implementing these effects in the renderer. However, the computational cost is large for rendering the training dataset. Instead, we propose using a heuristic augmentation, which is computationally efficient and flexible. Augmentations such as blurring and adding noise are commonly used in RGB-based methods. However, for polarized vision, our insight is that the augmentation \textbf{before} polarization signal processing (Eq. \ref{equ:four2stokes}, \ref{equ:stokes2dolpaolp}) is key to realism. This point, however, is less discussed in the previous learning-based SfP methods. Fig. \ref{fig:augmentation} demonstrates that applying augmentation before polarization signal processing creates a similar effect to real data.  

Given Stokes vectors rendered using an ideal pinhole camera model (the most common output for renderers), we first recover them into four linearly polarized images using the inverted version of Eq. \ref{equ:four2stokes}. Then we apply augmentation on four linearly polarized images. A Gaussian blur with random kernel size is first applied across four images to improve robustness to out‑of‑focus scenes. Subsequently, zero‑mean Gaussian noise is injected into each polarization image, with the noise strength randomly sampled. The rendered images are usually 16 or 32 bit \cite{jakob2022dr}, but the analog to digital converter of the polarization sensor \cite{SonyPolarSens} is just 12 bit. To fill this gap, we also add a quantization step converting the input images to 12 bit. Finally, we apply Eq. \ref{equ:four2stokes}, \ref{equ:stokes2dolpaolp} to get RGB, DoLP, and AoLP images. Note that the augmentation is only used for training.

\noindent \textbf{Network architecture and loss function}
We adopt a hybrid architecture that combines a UNet encoder–decoder with a frozen DINOv3 ConvNeXt backbone for surface normal estimation from polarization information. The model takes $s_0$ (RGB), $\mathrm{DoLP}$, and $\mathrm{AoLP}$ as input (computed according to Eqs. \ref{equ:four2stokes} and \ref{equ:stokes2dolpaolp}), and outputs a pixel-wise normal map. The UNet encoder processes all input channels. In parallel, only the RGB channels are fed into the DINOv3 branch, from which intermediate feature maps at four hierarchical stages are extracted. Feature fusion is performed in the decoder in a multi-scale manner. At each resolution level, the DINOv3 feature map is spatially aligned and concatenated with the corresponding UNet encoder feature, followed by an upsampling and convolutional block. Further architectural details are provided in the supplementary material. We supervise the predicted normal using cosine loss,
\begin{equation}
\mathcal{L} = \frac{1}{M} \sum_{i \in M}  1 - \mathbf{n}_i \cdot \hat{\mathbf{n}_i},
\label{equ:loss}
\end{equation}
where $M$ is the foreground region defined by the binary mask, $\mathbf{n}$ and $\hat{\mathbf{n}}$ are the ground-truth and predicted surface normal at pixel $i$.

\section{Training and Evaluation Datasets}
\subsection{Synthetic}
Given the insufficient realism and low diversity issues of existing SfP datasets, we construct a synthetic polarized dataset for training, termed \textbf{DTC-p}. We use 1,954 3D objects from the DTC dataset \cite{dong2025digital} for training and 40 objects for evaluation and testing, together with environment maps from Poly Haven \cite{polyhaven}, where 827 are used for training and 10 for evaluation and testing. All scenes are rendered using Mitsuba3 \cite{jakob2022dr} in polarized mode and a pBRDF model proposed by Baek \etal \cite{baek2018simultaneous}. Each scene randomly samples 1-10 objects and one environment map. Objects are randomly scaled and placed on a flat ground, and a simple overlap detection script is applied. Environment maps are randomly rotated. Camera positions are randomly sampled on a hemisphere around the scene. For each scene, we render Stokes vectors, along with an object mask and a ground-truth surface normal map with a resolution of 512 × 612 (height × width). In total, the dataset contains 40K training scenes, 1,000 validation scenes, and 1,000 test scenes. 

\subsection{Real}
\label{sec:data_real}
For real-world evaluation, we use both public datasets and datasets collected by us. Real-world evaluation for SfP has been challenging due to the scarcity of polarization datasets with ground-truth surface normals. Existing datasets are restricted to gray-scale images \cite{ba2020deep}, controlled lighting conditions \cite{lyu2023shape}, or objects with simple geometry \cite{lyu2024sfpuel, chen2024pisr, han2024nersp}.
To address these limitations, we capture a real-world evaluation dataset containing five objects with complex geometry, referred to as Our real \textit{w/ GT}. Following prior scanner-based approaches, we acquire ground-truth geometry using a high-end 3D scanner (EinScan Pro HD) and align the scanned meshes with captured images using Mitsuba3 \cite{jakob2022dr}. Due to the high cost of 3D scanning, this dataset is relatively small (5 scenes).
To complement it, we further construct a real-world polarization dataset without ground-truth normals for qualitative evaluation, named Our real \textit{w/o GT}. All polarization images are captured using a FLIR BFS-U3-51S5PC-C polarization camera equipped with a Sony IMX250MYR sensor \cite{SonyPolarSens}.
Overall, quantitative evaluation is performed on three datasets with ground-truth normals: PISR \cite{chen2024pisr}, SfPUEL \cite{lyu2024sfpuel}, and Our real \textit{w/ GT}. Qualitative evaluation is conducted on several public datasets \cite{dave2022pandora, li2024neisf, li2025neisf++, kurita2023simultaneous} and Our real \textit{w/o GT}.

\section{Experiments}
\subsection{Experimental setup}
\noindent \textbf{Implementation details} Our model is implemented in PyTorch. The UNet branch follows the architecture described in the original paper \cite{ronneberger2015u}, while the DINOv3 \cite{simeoni2025dinov3} branch adopts a ConvNeXt (base size) \cite{liu2022convnet} backbone. We use the Adam \cite{kingma2014adam} optimizer with an initial learning rate of 1e-4 and a batch size of 8. The learning rate is scheduled using StepLR, with a step size of 10 epochs and a decay factor of 0.5. Models are trained for 30 epochs on our DTC-p. The input training images are 512 × 612 (height × width). The DoLP and AoLP inputs are linearly mapped to [-1, 1], while RGB images are normalized using the ImageNet \cite{deng2009imagenet} mean and standard deviation. The training takes roughly two days on a single NVIDIA V100 GPU.

\noindent \textbf{Competitors} We choose the best SfP method SfPUEL \cite{lyu2024sfpuel}, generative VFMs StableNormal \cite{ye2024stablenormal}, Diffusion-E2E-FT \cite{garcia2025fine}, Discriminative VFM MoGe2 \cite{wang2025moge2}, and a commercial inverse rendering tool SwitchLight3 \cite{beeble_switchlight3} as our competitors. To better isolate the effects of the dataset and the model, we retrain \cite{garcia2025fine} on our proposed dataset, denoted as \cite{garcia2025fine} \textit{w/} DTC-p. Since this model only supports RGB inputs, we retrain it using only the RGB part of DTC-p.

\noindent \textbf{Evaluation metrics} Following existing single-shot normal estimation works \cite{bae2024rethinking}, we report the mean angular error (MAE) between the estimated and ground truth normals. The percentage of pixels with angular error below $11.25^\circ$ and $22.50^\circ$ is also calculated for a more comprehensive evaluation.

\begin{figure}[t]
  \centering
  \includegraphics[width=\linewidth]{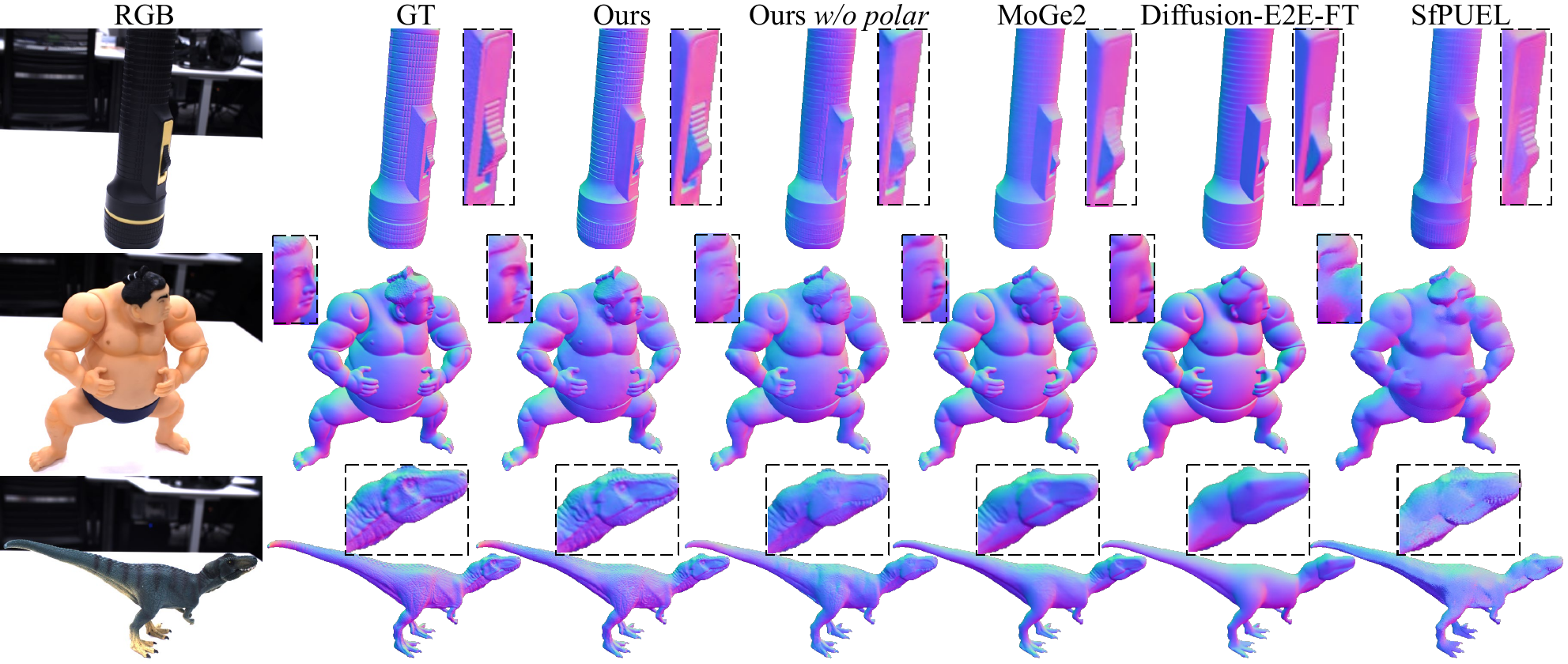}
  \caption{Qualitative comparisons on Our real \textit{w/ GT}.}
  \label{fig:compare}
\end{figure}

\begin{figure}[t]
  \centering
  \includegraphics[width=\linewidth]{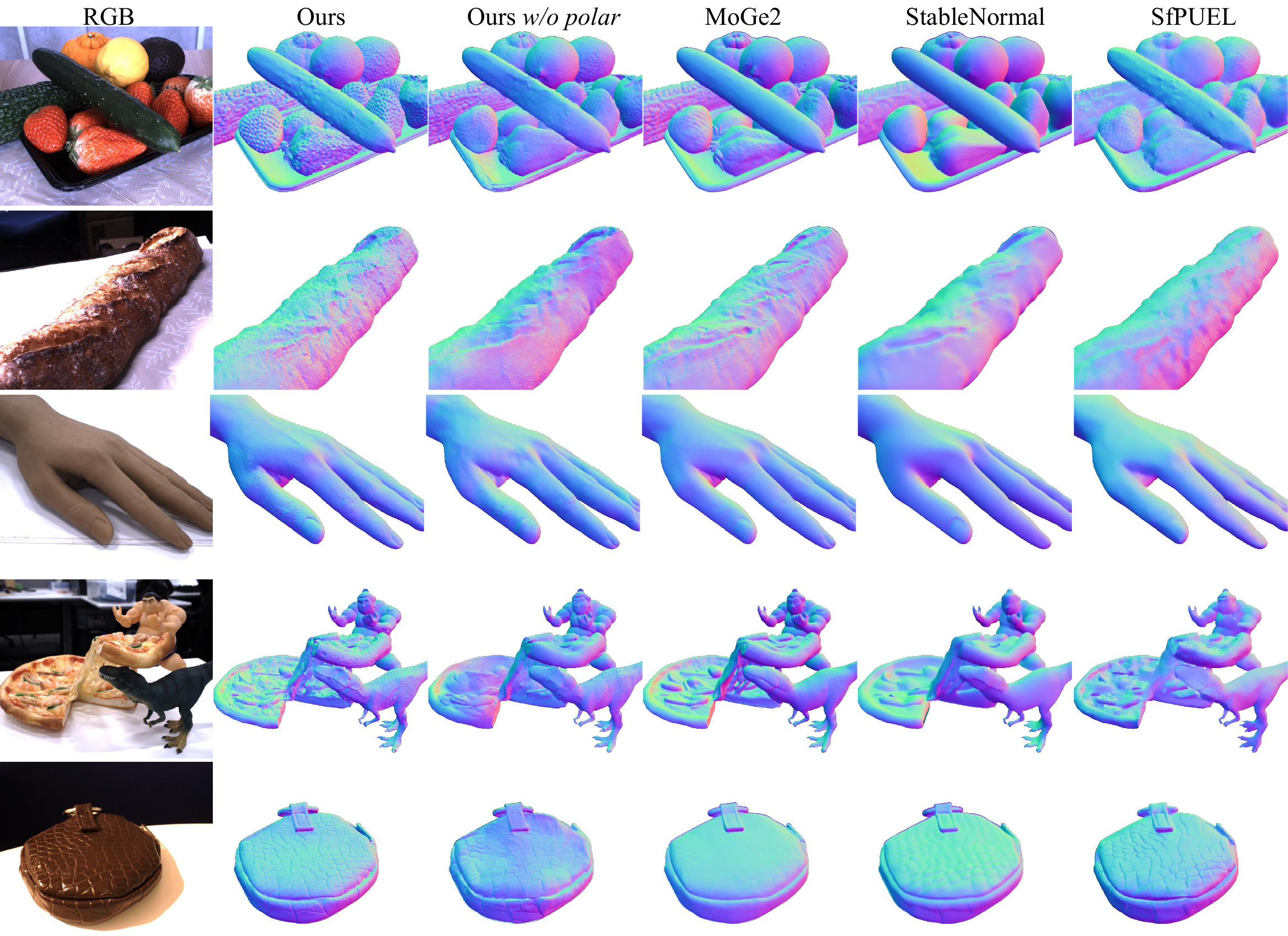}
  \caption{Qualitative comparisons on Our real \textit{w/o GT} (Please zoom in for details). }
  \label{fig:compare_no_gt}
\end{figure}

\begin{table}[t]
\centering
\small
\caption{Quantitative comparison on three real-world datasets. SwitchLight3 \cite{beeble_switchlight3} is a commercial tool and is not included in the ranking.}
\label{tab:ablation_comparison}
\resizebox{\linewidth}{!}{
\begin{tabular}{lccccccc}
\toprule
Method &
\multirow{2}{*}{\makecell{Training \\ Data}} &
\multicolumn{4}{c}{MAE $\downarrow$} &
\multicolumn{2}{c}{Acc $\uparrow$} \\
\cmidrule(lr){3-6}
\cmidrule(lr){7-8}
& 
& Our real \textit{w/ GT} & PISR \cite{chen2024pisr} & SfPUEL \cite{lyu2024sfpuel} & Avg
& $<11.25^\circ$ & $<22.50^\circ$ \\
\midrule
Ours & 40K & \textbf{12.63$^\circ$} & \textbf{12.32$^\circ$} & 12.76$^\circ$ & \textbf{12.54$^\circ$} & \textbf{58.2\%} & \textbf{88.5\%} \\
\textit{w/o polar} & 40K & 17.82$^\circ$ & 22.99$^\circ$ & 15.90$^\circ$ & 18.43$^\circ$ & 32.8\% & 70.9\% \\
\textit{w/o aug} & 40K & 14.26$^\circ$ & 15.93$^\circ$ & 13.88$^\circ$ & 14.55$^\circ$ & 48.2\% & 83.5\% \\
\textit{w/ post aug} & 40K & \underline{13.31$^\circ$} & \underline{15.31$^\circ$} & 13.31$^\circ$ & 13.84$^\circ$ & \underline{52.2\%} & \underline{85.4\%} \\
\textit{w/o DINO} & 40K & 14.83$^\circ$ & 15.33$^\circ$ & 14.99$^\circ$ & 15.03$^\circ$ & 45.2\% & 82.3\% \\
\textit{w/ SfPUEL data} & 20K & 14.58$^\circ$ & 16.93$^\circ$ & 12.38$^\circ$ & 14.33$^\circ$ & 49.4\% & 83.2\% \\
\midrule
MoGe2 \cite{wang2025moge2} & 8.9M &
14.46$^\circ$ & 16.73$^\circ$ & \textbf{10.88$^\circ$} & \underline{13.63$^\circ$} & 50.9\% & 85.3\% \\
StableNormal \cite{ye2024stablenormal} & 250K &
17.96$^\circ$ & 25.04$^\circ$ & 17.18$^\circ$ &
20.14$^\circ$ & 30.3\% & 67.5\% \\
SfPUEL \cite{lyu2024sfpuel} & 20K & 18.31$^\circ$ & 20.22$^\circ$ & \underline{11.16$^\circ$} & 15.96$^\circ$ & 44.0\% & 79.2\% \\
Diffusion-E2E-FT \cite{garcia2025fine} & 74K & 16.91$^\circ$ & 18.97$^\circ$ & 17.05$^\circ$ & 17.51$^\circ$ & 35.6\% & 73.2\% \\
\cite{garcia2025fine} \textit{w/} DTC-p & 40K & 16.46$^\circ$ & 16.74$^\circ$ & 11.85$^\circ$ & 14.69$^\circ$ & 50.5\% & 82.5\% \\

\midrule
\rowcolor{gray!20}
SwitchLight3 \cite{beeble_switchlight3} & -- &
14.32$^\circ$ & 16.66$^\circ$ & 9.36$^\circ$ &
12.96$^\circ $ & 55.4\% & 86.3\% \\
\bottomrule
\end{tabular}
}
\end{table}

\begin{figure}[t]
  \centering
  \includegraphics[width=\linewidth]{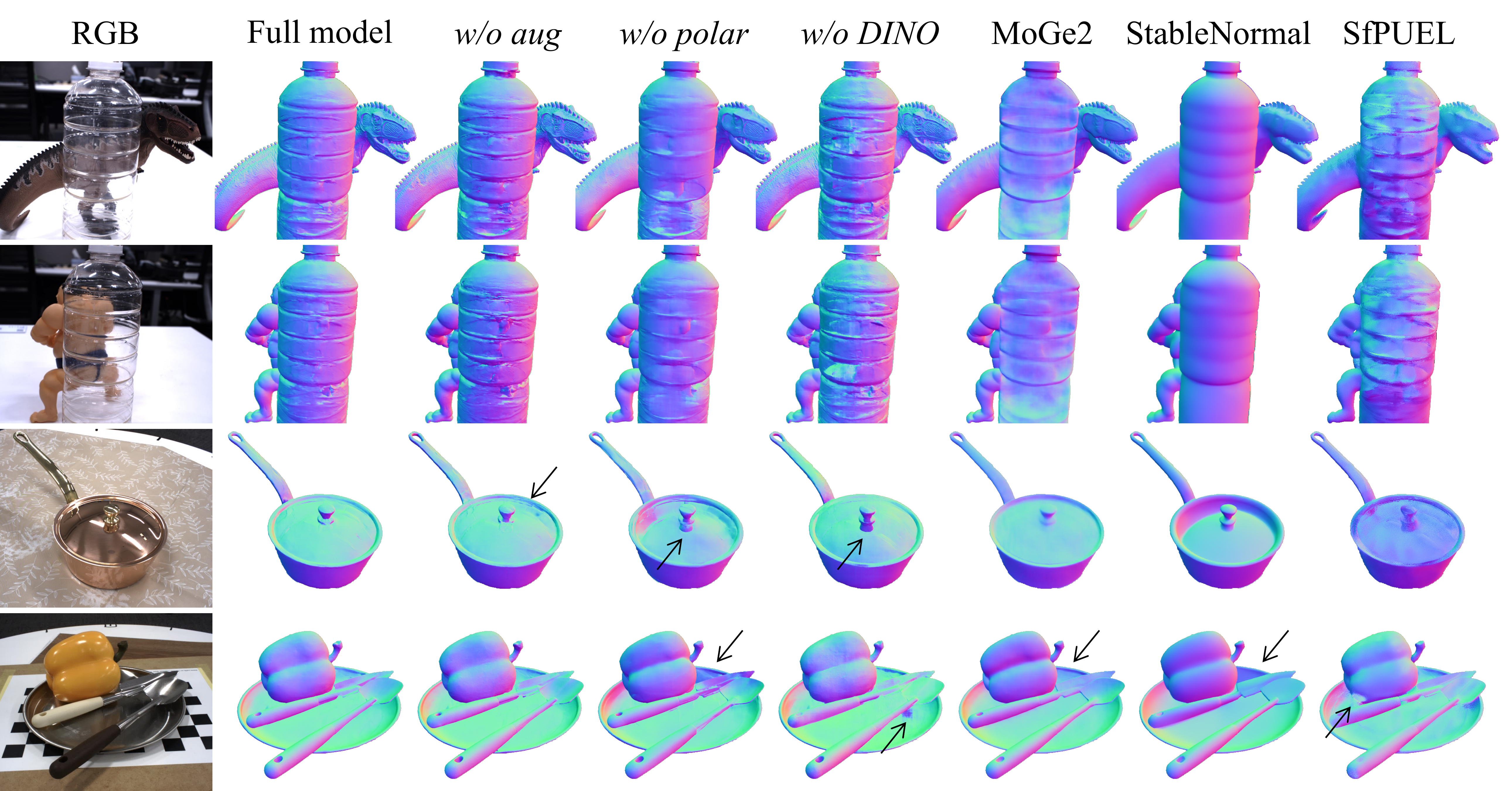}
  \caption{Out of distribution tests. Our training dataset does not contain transparent objects or conductors. Our full model is still robust to these unseen objects, while the ablated versions show various types of degradations. }
  \label{fig:ablation}
\end{figure}

\begin{figure}[t]
  \centering
  \includegraphics[width=\linewidth]{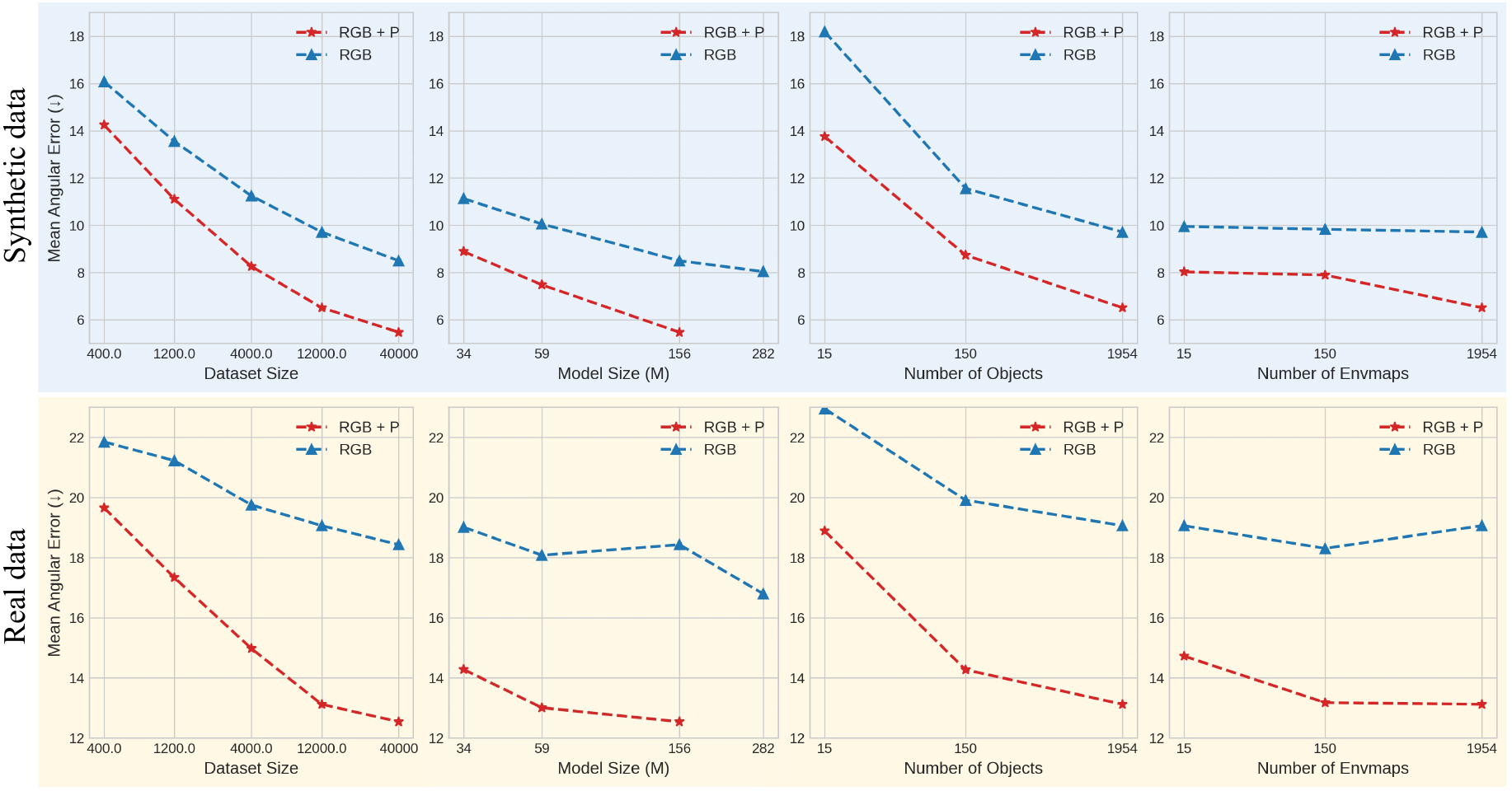}
  \caption{Ablation study of number of training scenes, model size, and number of objects and environment maps  used for rendering the training data. MAE of synthetic data are calculated on the test part of DTC-p. MAE of real data are calculated using three datasets. Log scaling is applied to horizontal axis for better visualization.}
  \label{fig:ablation_plot}
\end{figure}


\begin{figure}[t]
  \centering
  \includegraphics[width=\linewidth]{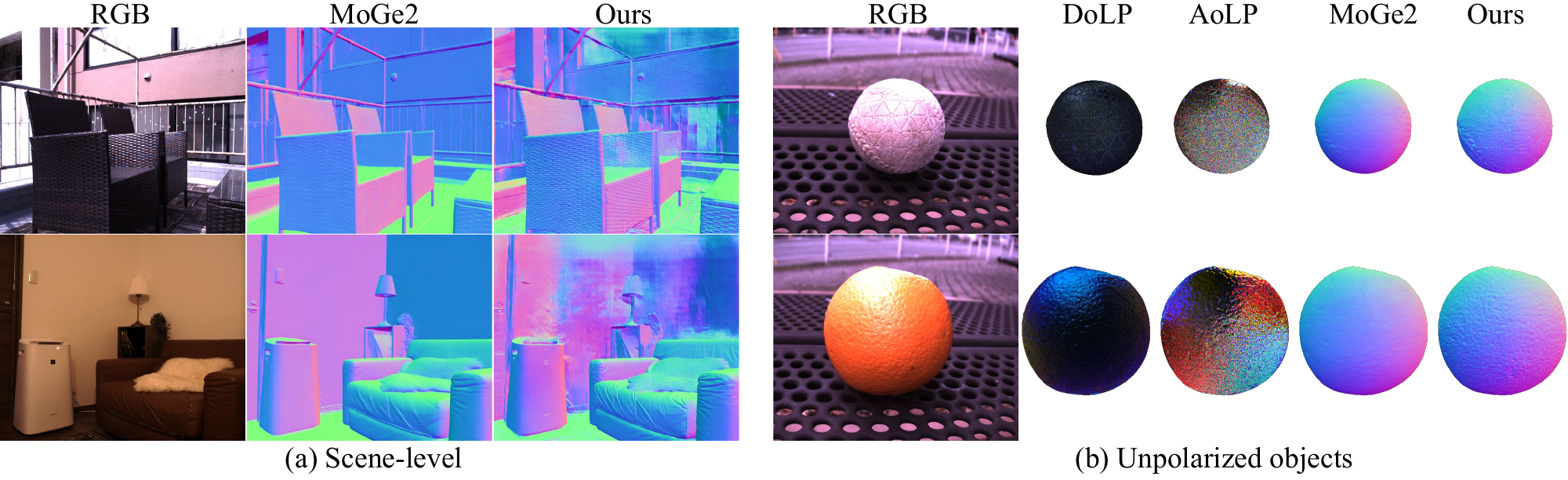}
  \caption{Failure cases. (a) Although the model recovers fine-grained geometry of individual objects, it fails to correctly understand background structures such as walls or buildings.
  (b) When the polarization signal is strong (bottom), the model produces high-fidelity geometry. However, when the target object is nearly unpolarized (top), the DoLP is close to zero and the AoLP becomes extremely noisy, showing no noticeable improvement over RGB-only VFMs.}
  \label{fig:f_cases}
\end{figure}

\subsection{Comparisons}
We report both qualitative comparisons in Fig. \ref{fig:teaser}, \ref{fig:compare}, 
\ref{fig:compare_no_gt}, \ref{fig:ablation} and quantitative comparisons in Tab. \ref{tab:ablation_comparison}. Our model consistently achieves a low MAE and high accuracy, yielding the best average performance across three datasets. The qualitative results further demonstrate the superiority of our model. Our method better recovers the detailed geometric structure, while RGB-only VFMs such as MoGe2 \cite{wang2025moge2}, Diffusion-E2E-FT \cite{garcia2025fine}, and StableNormal \cite{ye2024stablenormal} tend to recover an overly smooth surface. SfPUEL \cite{lyu2024sfpuel} also recovers details but suffers from issues like texture copying (see Fig. \ref{fig:teaser} for reference).

\subsection{Ablation study}
\subsubsection{Model ablation} To better understand our model, we compare it with several ablated variants by removing different components: polarization sensor-aware augmentation (\textit{w/o aug}), polarization cues (\textit{w/o polar}, \ie only RGB images are fed into the UNet branch), and DINOv3 features (\textit{w/o DINO}, \ie training a simple UNet). We further conduct an ablation study by moving the augmentation stage after polarization signal processing (\textit{w/ post aug}). We report the qualitative results in Fig. \ref{fig:ablation} and the quantitative results in Tab. \ref{tab:ablation_comparison}.  To study the effect of model size, we further replace the DINOv3 encoder (base) with other sizes and adjust the UNet channel width accordingly. The results are presented in Fig.~\ref{fig:ablation_plot}.

From the quantitative results, we observe that \textbf{polarization cues contribute the most to performance}, reducing MAE by 32.0\%. The next most important components are the DINOv3 prior (16.6\%) and data augmentation (13.8\%). The qualitative results match these findings, where clear degradations and artifacts appear when any of these components are removed. Moreover, \textbf{performing augmentation before polarization signal processing is crucial}. As shown in Tab.~\ref{tab:ablation_comparison}, pre-augmentation consistently outperforms post-augmentation across all datasets. More importantly, \textbf{polarization cues are robust to synthetic-to-real domain gaps}. As shown in Fig.~\ref{fig:ablation_plot}, for all experiments, the performance gap between models with and without polarization cues is consistently larger on real data than on synthetic data. From the model size ablation, we further reveal \textbf{the potential of polarization cues in reducing model size}. With polarization cues, even the smallest model with only 34M parameters outperforms the largest RGB-only model with 282M parameters on real-world evaluation. This result strongly suggests that polarization cues remain valuable in the era of VFMs. While VFMs provide strong performance, their large model size leads to high inference costs. By incorporating polarization cues, we can significantly reduce the model size while maintaining strong performance.
 
\subsubsection{Data ablation} 
 Most previous SfP works focus on model ablation, while dataset ablation has been explored much less, despite its importance for learning-based methods. We address this gap by conducting several dataset ablation studies.
 We first train our model using the SfPUEL dataset \cite{lyu2024sfpuel} (\textit{w/ SfPUEL data}) to demonstrate the importance of high-quality training data (Tab. \ref{tab:ablation_comparison}). We further evaluate performance degradation when reducing the number of training scenes, the number of objects, and the number of environment maps used in rendering. Due to limited computational resources, experiments involving object and environment map variations are conducted with 12K training scenes.

From these results, we find that \textbf{polarization cues help reduce the required training data size}. As shown in the real-world results in Fig.~\ref{fig:ablation_plot}, models with polarization cues achieve better performance than RGB-only models even when trained with 33$\times$ fewer scenes.
In addition, we observe that \textbf{object diversity is critical}. Reducing the number of objects used for rendering causes a significant performance drop, while reducing the number of environment maps has a much smaller impact. Furthermore, \textbf{object quality also plays an important role}. This is evidenced by comparing Ours \textit{w/ SfPUEL data} (Tab. \ref{tab:ablation_comparison}, row 5) with our model trained on DTC-p rendered using only 150 objects. Although the former uses 244 objects and 20K scenes, it has a larger MAE (14.33$^\circ$) than the latter (14.27$^\circ$), which uses only 150 objects and 12K scenes. This result indicates that the realism of objects is also important in addition to dataset scale.
 
\subsection{Out-of-Distribution Test and Failure Cases}
To demonstrate the robustness of our model, we evaluate it on several out-of-distribution objects that are not seen during training (Fig. \ref{fig:ablation}), including transparent objects or conductors (our training data are rendered using a dielectric pBRDF). In addition, we test our model under two extreme conditions (Fig.~\ref{fig:f_cases}): scene-level surface normal estimation and nearly unpolarized objects.

The experimental results show that our model generalizes well to out-of-distribution data. For objects such as conductors, our method even outperforms RGB-only VFMs. However, for scene-level normal estimation, although our model predicts high-quality normals for individual objects, it lacks global scene understanding. As a result, incorrect normals are estimated for background walls and buildings. This behavior is expected since our model is trained only on object-level data.
Moreover, our method does not outperform RGB-only VFMs when polarization cues are severely degraded. In cases where the AoLP is dominated by noise, polarization information becomes unreliable. For example, fuzzy and white objects, such as the baseball shown in Fig.~\ref{fig:f_cases}, are nearly unpolarized due to dominant diffuse reflection and multiple inter-reflections between micro-surfaces. In addition, the polarization camera is only 12-bit, which limits its ability to capture subtle intensity differences among the four polarized images, resulting in a noisy AoLP signal.


\section{Limitations and Future Works}
\subsubsection{Target scene and material}
The current method works at the object level and only supports opaque and dielectric materials. A more general approach should handle both object-level and scene-level normal estimation while accommodating challenging materials like conductors and transparent materials. Although out-of-distribution tests have shown robustness to unseen content, there is still room for performance improvement. Expanding the method to include these features could lead to exciting applications. We believe this can be achieved by simply incorporating such scenes into the training data.


\subsubsection{Robustness to nearly unpolarized objects}
In Fig. \ref{fig:f_cases}, we present failure cases where the polarization images exhibit strong noise. This problem occurs when the target is nearly unpolarized. In such cases, the polarization sensor’s dynamic range is insufficient to capture such subtle polarization signals, causing the AoLP to be dominated by noise. Exploring methods to reliably capture weak polarization signals is an interesting direction for future work.

\subsubsection{Exploring advanced fusion architecture}
The approach of fusing multi-modal sensors is still an active research direction \cite{li2025stitchfusion}. As the first attempt of incorporating polarization cues into DINOv3 encoders, we only apply a simple multi-layer concatenation architecture. However, we believe there is substantial room for improvement in the fusion method and it is worth further exploration.

\section{Conclusions}



In this work, we proposed an SfP model that outperforms both existing SfP methods and RGB-only VFMs for object-level normal estimation. We show that the key to high accuracy lies in data realism and polarization sensor-aware augmentation. In addition, incorporating DINOv3 priors significantly improves robustness to unseen objects. Notably, these gains are achieved without relying on elaborate network designs or specialized training tricks; instead, a straightforward end-to-end pipeline is sufficient when polarization cues are properly modeled.  Beyond strong results, our study reveals that polarization cues are effective in bridging the synthetic-to-real gap and improving data and parameter efficiency. These findings demonstrate that polarization remains an efficient cue in the era of vision foundation models. We hope this study encourages renewed attention and inspires future research on physics-based sensing modalities.
%
%
\bibliographystyle{splncs04}
\bibliography{main}
\end{document}